VITORINO RAMOS[Π]
aLife Art Architecture Lab
+ CVRM / IST Technical Univ. of Lisbon

# On the Implicit and on the Artificial[Å]

## Morphogenesis and Emergent Aesthetics in Autonomous Collective Systems

"[…]QUESTION_HUMAN > If Control's control is absolute, why does Control need to control?
ANSWER_CONTROL > Control…, needs time.
QUESTION_HUMAN > Is Control controlled by his need to control?
ANSWER_CONTROL > Yes.
QUESTION_HUMAN > Why is Control need Humans, has you call them?
ANSWER_CONTROL > Wait! Wait…! Time are lending me…;
Death needs time like a Junkie… needs Junk.
QUESTION_HUMAN > And what does Death need time for?
ANSWER_CONTROL > The answer is so simple! Death needs time for what it kills to grow in! […]",

in Dead City Radio, *William S. Burroughs / John Cale*, 1990.

**Imagine a "machine" where there is no pre-commitment to any particular representational scheme: the desired behaviour is distributed and roughly specified simultaneously among many parts, but there is minimal specification of the mechanism required to generate that behaviour, i.e. the global behaviour evolves from the many relations of multiple simple behaviours. A machine that lives to and from/with Synergy. An artificial super-organism that avoids specific constraints and emerges within multiple low-level implicit bio-inspired mechanisms.**

The emergence of complex behaviour in any system consisting of interacting simple elements is among the most fascinating phenomena of our world. Examples can be found in almost every field of today's scientific interest, ranging from coherent pattern formation in physical and chemical systems, to the motion of swarms of animals in biology, and the behaviour of social groups. In the life and social sciences, one is usually convinced that the evolution of social systems is determined by numerous factors, difficult to grasp, such as cultural, sociological, economic, political, ecological, etc. However, in recent years, the development of the interdisciplinary fields "science of complexity", along with "artificial life" (*aLife*), has lead to the insight, that complex dynamic processes may also result from simple interactions. Moreover, at a certain level of abstraction, one can also find many common features between complex structures in very different fields.

*Francis Heylighen*, mentor of the Principia Cybernetic Project, an international organization (PCP, http://pespmc1.vub.ac.be/) points precisely to this paradigm shift, with a remarkable historical perspective, namely in what concerns the view within the social sciences, using biology as a metaphor, and more recently those from complexity science. In "The Global Superorganism: an Evolutionary-Cybernetic Model of the Emerging Network Society" (*Journal of Social and Evolutionary Systems*, 2001) he writes:

---

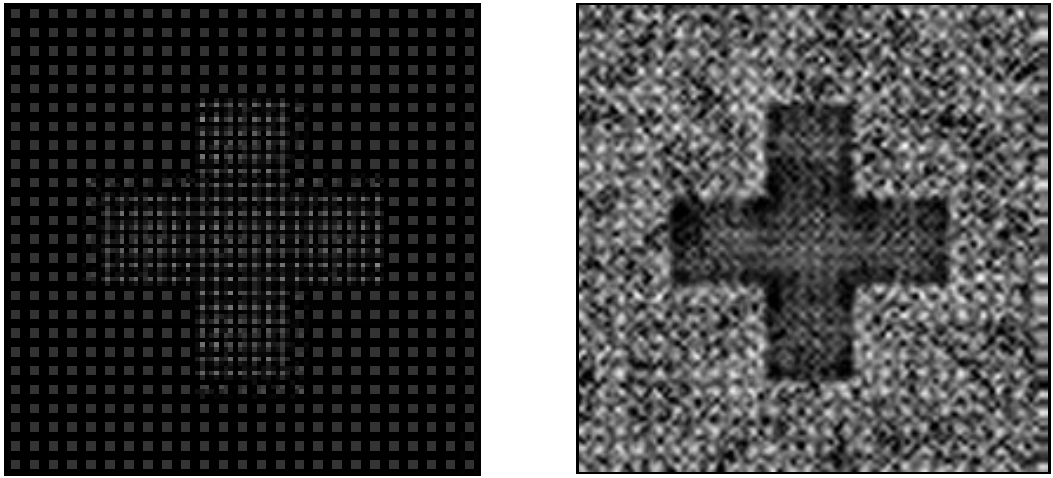

On the right, a synthetic test image composed by small squares of different sizes, but perceived by us (humans) as a big cross, probably due to our inner perceptual grouping *Gestalt* laws (*Wertheimer*, 1910). On the left, the perceived cross by an artificial ant colony as a function of the spatial distribution of pheromone (Swarm Cognitive Map), at *T*=1000 (*Ramos*, 1998-2000).

[…] It is an old idea that society is in a number of respects similar to an organism, a living system with its cells, metabolic circuits and systems. As an example, the army functions like an immune system, protecting the organism from invaders, while the government functions like the brain, steering the whole and making decisions. In this metaphor, different organizations or institutions play the role of organs, each fulfilling its particular function in keeping the system alive, an idea that can be traced back at least as far as *Aristotle*, being a major inspiration for the founding fathers of sociology, such as *Comte*, *Durkheim* and especially *Spencer* […]

Then at this point, *Heylighen* stresses the importance of recognizing the underlying component of complexity in nature, a bottom-up view common to the field of Artificial Life:

[…] The organismic view of society has much less appeal to contemporary theorists. Their models of society are much more interactive, open-ended, and non-deterministic than those of earlier sociologists, and they have learned to recognize the intrinsic complexity and unpredictability of society. The static, centralized, hierarchical structure with its rigid division of labor that seems to underlie the older organismic models appears poorly suited for understanding the intricacies of our fast-evolving society. Moreover, a vision of society where individuals are merely little cells subordinated to a collective system has unpleasant connotations to the totalitarian states created in the last century. As a result, the organismic model is at present generally discredited in sociology […]

Similarly, biology has traditionally started at the top, viewing a living organism as a complex biochemical machine, and has worked *analytically* down from there through the hierarchy of biological organization – decomposing a living organism into organs, tissues, cells, organelles, and finally molecules – in its pursuit of the mechanisms of life. Analysis means 'the separation of an intellectual or substantial whole into constituents for individual study' (that is, by top-down reductionist approaches). By composing our individual understandings of the dissected component parts of living organisms, traditional biology has provided us with a broad picture of the mechanics of life on Earth.
In the meantime, however, new scientific developments have done away with rigid, mechanistic views of organisms (*Heylighen*). As pointed by *Langton*, there is more to life than mechanics – there is also dynamics. Life depends critically on principles of dynamical self-organization that have remained largely untouched by traditional analytic methods. There is a simple explanation for this – these self-organized dynamics are fundamentally non-linear phenomena, and non-linear phenomena in general depend critically on the interactions between parts: they necessarily disappear when parts are treated in isolation from one another, which is the basis for any analytic method. Rather, non-linear phenomena are most appropriately treated by a *synthetic* approach, where synthesis means "the combining of separate elements or substances to form a coherent whole'. In non-linear systems, the parts must be treated in each other's presence, rather than independently from one another, because they behave very differently in each other's presence than we would expect from a study of the parts in isolation. As suggested by *Langton*, the key concept in *aLife* is emergent behaviour. Natural life emerges out of the organised interactions of a great number of nonliving molecules, with no global controller responsible for the behaviour of every part. Rather, every part is a behaviour itself, and life is the behaviour that emerges from out of all of the local interactions among individual behaviours. It is this bottom-up, distributed local determination that aLife employs in its primary methodological approach to the generation of lifelike behaviours. Of course, there is no universally agreed definition of life. The concept covers a cluster of properties, most of which are themselves philosophically problematic: self-organization, emergence, autonomy, growth, development, reproduction, evolution, adaptation, responsiveness, and metabolism. Scientists differ about the relative importance of these properties, although it is generally agreed that the possession of most (not necessarily all) of them suffices for something to be regarded as alive. One common concept however is shared by all: complex behaviour can emerge from a system consisting of interacting simple elements. Once more, *Heylighen* stresses biology as an example :

> […] when studying living systems, biologists no longer focus nowadays on the static structures of their anatomy, but on the multitude of interacting processes that allow the organism to adapt to an ever changing environment. Recently, the variety of ideas and methods that is commonly grouped under the head of *the sciences of complexity*, has led to understanding that artificial organisms can be self-organizing, adaptive systems. Most processes in such systems are decentralized, non-deterministic and in constant flux. They thrive on noise, chaos and creativity. Their collective intelligence emerges out of the free interactions between individually autonomous components […]

In fact, as I see it, those processes should be viewed as behaving like a swarm. Rather than take living things apart, Artificial Life attempts to put living things together within a *bottom-up* approach, that is, beyond *life-as-we-know-it* into the realm of *life-as-it-could-be* (*Langton*), generating *lifelike* behaviour, and focusing on the problem of creating behaviour generators, inspired on the nature itself, even if the results (what emerges from the process) have no analogues in the *natural* world. The key insight into the natural method of behaviour generation is gained by noting that nature is fundamentally parallel. This is reflected in the "architecture" of natural living organisms, which consist of many millions of parts, each one of which has its own behavioural repertoire. As we know, living systems are highly distributed and quite massively parallel.

The so-called *nouvelle* Artificial Intelligence (AI) and aLife are each concerned with the application of computers to the study of complex, natural phenomena. Apart from traditional and symbolic hard-specific top-down AI in the sixties and seventies, both are nowadays concerned with generating complex behaviour, in a bottom-up manner, turning their attention from the *mechanics* of phenomena to the *logic* of it. The first computational approach to the generation of lifelike behaviour was due to the mathematician *John Von Neumann*. In the words of his colleague *Arthur W. Burks*, *Von Neumann* was interested in the general question:

> […] What kind of logical organization is sufficient for an automaton to reproduce itself ? This question is not precise and admits to trivial versions as well as interesting ones. *Von Neumann* had the familiar natural phenomenon of self-reproduction in mind when he posed it, but he was trying to simulate the self-reproduction of a natural system at the level of genetics and biochemistry. He wished to abstract from the natural self-reproduction problem its logical form […]

This approach is probably the first to capture the essence of Artificial Life (replace, for instance, references to 'self-reproduction' in the above with references to any other biological phenomena). From this "kinematic model" of *Von Neumann*, a genuine self-reproduction mechanism implemented in the sixties, *Stan Ulam* suggested an appropriate formalism where the logical form of the process is completely distinguish from the material counterpart, which has come to be know as a *Cellular Automata* (CA). In brief, a CA consists of a regular lattice of (many) *finite automata*, which are the simplest formal models of machines. A finite automata can be in only one of a finite number of states at any given time, and its transition between states from one time-step to the next are governed by a *state-transition table*: given a certain input and a certain internal state, the state-transition table specifies the state to be adopted by the finite automata at the next time step. In a CA, the necessary input is derived from the states of the automata at neighbouring lattice-points. Thus the state of an cellular automata at time $t+1$ is a function of the states of the automata itself and its immediate neighbours at time $t$. All the finite automata in the lattice (group of cells) obey the same transition-table (rule table) and every cell changes his state at the same instant, time-step after time-step. CA's are a good example of the kind of computational paradigm sought after by Artificial Life: bottom-up, parallel, local determination of behaviour with minimal specification, and emerging complex phenomena from simple rules.

In order to study any natural phenomena, scientists are turning to a separation. A need to separate the notion of a formal specification of a machine (any that will reproduce the phenomena itself) – that is, a specification of the *logical structure* of the machine – from the notion of a formal specification of a machines's behaviour – that is, a specification of transitions that the machine will undergo. In general, we cannot derive behaviours from structure, nor can we derive structure from behaviours. So instead, in order to determine the behaviour of some machines and coupled phenomena, there is no recourse but to run them and see how they behave. This has consequences for the methods by which we (or nature) go about *generating* behaviour generators themselves, and from which any evolutionary and adaptive process seems to be essential. As an illustration, the most salient characteristic of living systems, from the behaviour generation point of view, is the *genotype/phenotype* distinction. The distinction is essentially one between a specification of machinery – the *genotype* – and the behaviour of that machinery – the *phenotype*.

The *genotype* is the complete set of genetic instructions encoded in the linear sequence of nucleotide bases that makes an organism's DNA. The *phenotype* is the physical organism itself – the structures that emerge in space and time as the result of the interpretation of the genotype on a particular environment. The process by which the phenotype develops through time under the direction of the genotype is called *morphogenesis*. Simulation plays an essential role in the study of *morphogenesis*. This was anticipated as early as 1952 by *Turing*, who wrote:

[…] The difficulties are such that one cannot hope to have any very embracing *theory* of such processes, beyond the statement of equations. It might be possible, however, to treat a few particular cases in detail with the aid of a digital computer. This method has the advantage that it is not so necessary to make simplifying assumptions as it is when doing a more theoretical type of analysis […]

What is notable is that these 1952 *Turing* words appears to have already the embedded features that characterise *bottom-up* approaches, in detriment of other kinds of approaches strictly *reductionist* (e.g. *top-down*). As an aside evidence, note the last *Turing* words on this sentence: it is not so necessary to make simplifying assumptions as it is when doing a more theoretical type of analysis […]. Visualisation itself, of simulation results facilitates their interpretation, and is used as a method for evaluating models. Lacking a formal measure of what makes two patterns or forms (such as trees) look alike (task that is, as we known, mainly related to the idea of perception), we rely on visual inspection comparing the models with the reality. Important however in these models, is that the natural and synthetic pigmentation patterns differ in details, yet we perceive them as fairly similar or familiar.

In *morphogenesis*, the individual genetic instructions are called *genes* and consist of short stretches of DNA. These instructions are *executed* (expressed) when their DNA sequence is used as a template for transcription. One may consider the genotype as a largely unordered 'bag' of instructions (a rule table, an alphabet, a group of primitives), each one of which is essentially the specification for a *machine* of some sort – passive or active. When instantiated, each such machine will enter into ongoing logical mechanisms, consisting largely of local interactions between other such machines. Each such instruction will be *executed* when its own triggering conditions are met and will have specific, local effects on structures in other cells (their neighbors). Furthermore, each such instruction will operate within the context of all the other instructions that have been – or are being – executed.

The phenotype, then, consists of the structures and dynamics that emerge through time in the course of the execution of the parallel, distributed computation controlled by this genetic *bag* of instructions. Since genes interactions with one another are highly non-linear, the phenotype is a non-linear function of the genotype. As mentioned briefly above, the distinction between linear and non-linear systems is fundamental, and provides excellent insight into why the principles underlying the dynamics of life (or many other natural phenomena) should be so hard to find and understand. The simplest way to state the distinction is to say that linear systems are those for which the behaviour of the whole is just the sum of the behaviour of its parts, while for non-linear systems, the behaviour of the whole is more than the sum of its parts. Linear systems are those which obey the *principle of superposition*. We can break up complicated linear systems into simpler constituents parts, and analyse these parts *independently*. Once we have reached an understanding of the parts in isolation, we can achieve a full understanding of the whole system by *composing* our understandings of the isolated parts. This is the key feature of linear systems: by studying the parts in isolation we can learn everything we need to know about the complete system. Nature, however, is generally non-linear, where this type of approach is often impossible. Non-linear systems do *not* obey the principle of superposition. Even if we could break such systems up into simpler constituents parts, and even if we could reach a complete understanding of the parts in isolation, we would not be able to compose our understandings of the individual parts into an understanding of the whole system. The key feature of non-linear systems is that their primary behaviours of interest are properties of the interactions between parts, rather than being properties of the parts themselves, and these interaction-based properties necessarily disappear when the parts are studied independently. Analysis has not proved anywhere near as effective when applied to non-linear systems: the non-linear system must be treated as a whole. A different approach to the study of non-linear systems involves the inverse of analysis: *synthesis*. Rather than start with the behaviour of interest and attempting to analyse it into its constituent parts, we should start with constituent parts and put them together in the attempt to synthesize the behaviour of interest. Life, in the same way, is a property of *form*, not *matter*, a result of organization and re-organization of matter rather than something that inheres in the matter itself. Neither nucleotides nor amino acids nor any other carbon-chain molecule is alive – yet put them together in the right way, and the dynamic that emerges out of their interactions is what we call life. It is effects, not things, upon which life is based – life is a kind of behaviour, not a kind of stuff – and as such, it is constituted of simpler behaviours, not simpler stuff. Behaviours themselves can constitute the fundamental parts of non-linear systems – *virtual parts*, which depend on non-linear interactions between physical parts for their very existence. Isolate the physical parts and the virtual parts cease to exist. It is the virtual parts of living systems that Artificial Life is after, and synthesis is its primary methodological tool.

Computers, provide (and should be viewed as) as an important laboratory tool for the study of life and many natural phenomena, as an alternative devoted exclusively to the incubation of information structures. The advantage of working with information structures is that information has no intrinsic size. The computer is the *tool* for the manipulation of information, whether that manipulation is a consequence of our actions or a consequence of the actions of the information structure themselves. Computers themselves will not be alive, rather they will support informational universes within which dynamic populations of informational 'molecules' (or *memes*, as proposed by *Dawkins*, as the cultural information genes, or vehicles, within one specific society) engage in informational 'biochemistry'. This view of computers as workstations for performing scientific experiments within artificial universes is fairly new, but is rapidly becoming accepted as a legitimate, even necessary, way of pursuing science. In the days before computers, scientists worked primarily with systems whose defining equations could be solved analytically, and ignored those whose defining equations could *not*. This was the case, for instance, in many analytical systems trying to explain how the global weather changes, or trying to forecast the behaviour of a fire propagating in a specific terrain. As we now know, global weather is a chaotic non-linear system, where a flap of a butterfly wing in Peking can develop a huge storm in New York, few days later. In the absence of analytical possible solutions, the equations would have to be integrated over and over again, essentially simulating the time behaviour of the system. Without computers to handle the mundane details of these calculations, such an undertaking was unthinkable except for the simplest cases. Given these mundane calculations to computers, the realm of numerical simulation is opened up for exploration. 'Exploration' is an appropriate term for the process, because the numerical simulation of systems allows one to explore the system's behaviour under a wide range of parameter settings and initial conditions. The heuristic value of this experimentation cannot be overestimated. One often gains tremendous insight for the essential dynamics of a system by observing its behaviour under a wide range of initial conditions. Moreover, computers are beginning to provide scientists with a new paradigm for modeling the world. When dealing with essentially unsolvable governing equations, the primary reason for producing a formal mathematical model (the hope of reaching an analytic solution by symbolic manipulation) is lost. It has become possible, for example, to model turbulent flow in a fluid by simulating the motions of its constituent particles – not just approximating *changes* in concentrations of particles at particular points, but actually computing their motions exactly. The same is true for understanding and modeling people in overcrowded soccer stadiums, or for instance, in gaining insight on how traffic jams emerge, from very simple inner rules. Again, the best way to tackle it, is to look at the whole process, synthesizing which basic and simple logical rules (generally independent from the phenomena itself) govern the multitude of parts, emerging a global and complex behaviour. What is essential in these types of systems, is not the parts and their intrinsic natures (at least strictly), but namely their relationships (among themselves and with their environment).

Within this same context, let us return again to the genotype/phenotype distinction and on the possibility of the development of a behavioural phenotype. One paradigmatic model is the one of *Craig Reynolds*, who in 1987 has implemented a simulation of flocking behaviour. Now, if we think for a moment, none type of analytical differential equations was been able to tackle (or model) this type of natural phenomena. In the *Reynolds* model, however – which is meant to be a general platform for studying the qualitatively similar phenomena of flocking, herding and schooling – one has a large collection of autonomous but interacting objects (which *Reynolds* refer as *Boids*), inhabiting a common simulated environment.

The modeler can specify the manner in which the individual *Boids* will respond to *local* events or conditions. The global behaviour of the aggregate of *Boids* is strictly an emergent phenomena, where none of the rules for the individual *Boids* depends on global information, and the only updating of the global state is done on the basis of individual *Boids* responding to local conditions. Note that, the underlying system *nature* is similar in many ways to a *Cellular Automata*, mentioned earlier. Again, each *Boid* (*cell* for the CA) in the aggregate shares the same behavioural 'tendencies':

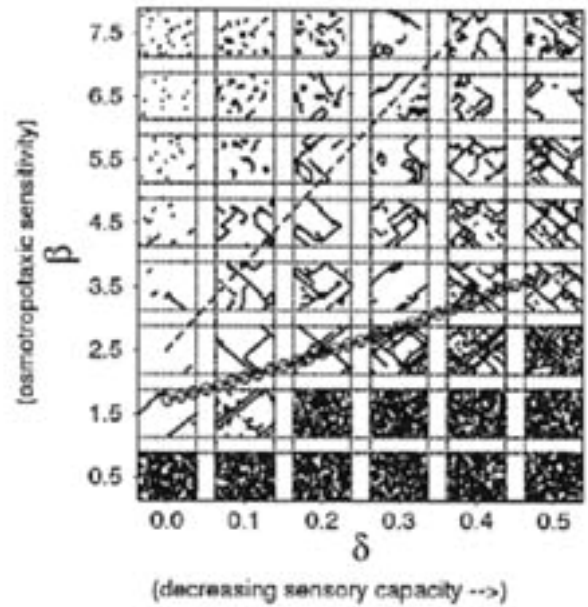

**SWARM COGNITIVE MAP MODEL:** The present paper is based on Chialvo and Millonas work [1], which is probably, one of the simplest (local, memoryless, homogeneous and isotropic) models which leads to trail forming, where the formation of trails and networks of ant traffic are not imposed by any special boundary conditions, lattice topology, or additional behavioural rules.

The parameter β is associated with the osmotropotaxic sensitivity. Osmotropotaxis, is related to a kind of instantaneous pheromonal gradient following. In practical terms, this parameter controls the degree of randomness with wich each ant follows the gradient of pheromone. For low values of β the pheromone concentration does not greatly affect its choice, while high values cause it to follow pheromone gradient with more certainty.
On the other hand, 1/δ is the sensory capacity, which describes the fact that each ant's ability to sense pheromone decreases somewhat at high concentrations.

The state of an individual ant can be expressed by its position $r$, and orientation θ. Since the response at a given time is assumed to be independent of the previous history of the individual, it is sufficient to specify a transition probability from one place and orientation $(r,\theta)$ to the next $(r',\theta')$ an instant later. The response function can effectively be translated into a two-parameter transition rule between the cells by use of a pheromone weigthing function:

$$W(\sigma) = \left(1 + \frac{\sigma}{1+\delta\sigma}\right)^\beta \quad \text{Eq. (1)}$$

This equation measures the relative probabilities of moving to a cite $r$ (in our context, to a pixel) with pheromone density σ(r).

Figure 1 - Effect of β and δ on the pheromonal map. Physiological phase plot produced by Chialvo and Millonas in [1], after 1000 time steps starting from a random initial distribution.

· To maintain a minimum distance from other objects in the environment, including other *Boids*,
· To match velocities with *Boids* in its neighbourhood, and
· To move towards the perceived centre of mass of the *Boids* in its neighbourhood.

These are the only rules governing the behaviour of the aggregate. These rules, then, constitute the generalized genotype of the *Boids* system. What is amazing, is that they say nothing about structure, or growth and development, or even about birds nature, but they determine the behaviour of a set of interacting autonomous objects, out of which very natural motion emerges. With the right settings for the parameters of the system, a collection of *Boids* released at random positions within a volume will collect into a dynamic flock, which flies around environmental obstacles in a very fluid and natural manner, occasionally breaking up into sub-flocks as the flock flows around both sides of an obstacle. Once broken up into sub-flocks, the sub-flocks reorganize around their own, now distinct and isolated centre of mass, only to re-emerge into a single flock again when both (or more) sub-flocks emerge at the fair side of the obstacle and each sub-flock *feels* anew the mass of the other sub-flock.

**PROBABILISTIC DIRECTIONAL BIAS:** In addition to the former equation, there is a weigthing factor $w(\Delta\theta)$, where $\Delta\theta$ is the change in direction at each time step, i.e. measures the magnitude of the difference in orientation. This weighting factor ensures that very sharp turns are much less likely than turns through smaller angles. Example - One ant is coming from north:

$w(0) = 1$
$w(1) = 1/2$
$w(2) = 1/4$
$w(3) = 1/12$
$w(4) = 1/20$

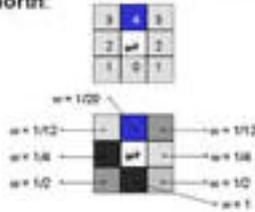

**Other examples** - Coming from east, southwest and northeast.

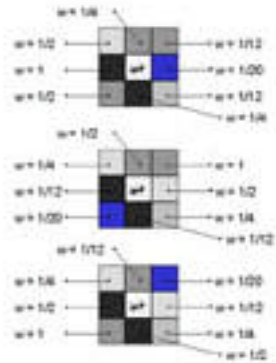

**NORMALISED TRANSITION PROBABILITIES:** Then, the normalised transition probabilities on the lattice to go from cell $k$ to cell $i$ are given by:

$$P_{ik} = \frac{W(\sigma_i)w(\Delta_i)}{\sum_{j/k} W(\sigma_j)w(\Delta_j)} \quad \text{Eq. (2)}$$

where the notation $j/k$ indicates the sum over all the pixels $j$ which are in the local neighbourhood of $k$. $\Delta i$ measures the magnitude of the difference in orientation for the previous direction at time $t-1$.

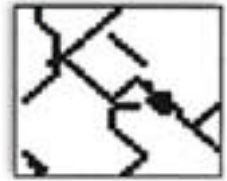

**EXTENDED MODEL TO GREY LEVEL HABITATS:** In order to model each ant perception of heterogeneous areas on the digital image habitat, robust metrics should be introduced. These metrics operate at local neighbourhoods for each ant in the colony, and in some sense they represent the individual "probability matrices" defined by Wilson, in our understanding of the mass behaviour. Then, by pheromone deposition, and extending this deposition to be proportional to those correlation values, each individual contributes to the swarm (whole) perception of the image, which in turn also serves as their habitat.

**1st EXTENSION:** Instead of constant pheromone deposition rate, a term not constant is included ($\Delta_b$):

$$T = \eta + p\Delta_b \quad \text{Eq. (3)}$$

Where $\Delta_b$ gives a measure of similarity between two different lattice windows, in terms of grey level spatial arrangement, i.e. $0 <= \Delta_b <= 1$ (Matching Properties), and $p$ is a constant.

**2nd EXTENSION:** This second extension is related with the metric used for correlation between two different lattice windows. Two alternative methods were used. The first one uses an evolutionary measure for image matching proposed by Bhat [2], which is based on the Ulam's distance - a well know ordinal measure from molecular biology, based on an evolutionary distance metric that is used for comparing real DNA strings. Given two strings, the Ulam's distance is the smallest number of mutations, insertions, and deletions that can be made within both strings such that the resulting substrings are identical. Then, Bhat reinterprets the Ulam's distance with respect to permutations that represent windows intensities expressed on an ordinal scale. The motivation for him to use this measure is twofold: it not only gives a robust measure of correlation between windows but also helps in identifying pixels that contribute to the agreement (or disagreement) between the windows.

The second one, which was also used for alternative experimental purposes is based on simple statistical measures as can be seen in equation (4):

$$\Delta_b = \left[ a\frac{|m_1 - m_2|}{Max|m_1 - m_2|} + b\frac{|\sigma_1^2 - \sigma_2^2|}{Max|\sigma_1^2 - \sigma_2^2|} + c\frac{S}{S_{max}} \right](a+b+c)^{-1} \quad \text{Eq. (4)}$$

The first term, computed through differences in simple averages is responsible for finding differences on grey level overall intensity values, while the second measures differences on windows grey level homogeneity values through variance computations. The last term measures successful matching properties between windows considering all types of permutations (where $(a+b+c)=1$; when comparing two windows, grey level average intensities in window one are represented by $m_1$, while $\sigma_1^2$ represents the variance for the same window).

 

The flocking behaviour itself constitutes the generalized phenotype of the *Boids* system. It bears the same relation to the genotype as an organism's morphological phenotype bears to its molecular genotype. The same distinction, between the specification of machinery and the behaviour of machinery is evident. Through development (or time), local rules governing simple non-linear interactions at the lowest level of complexity emerge global behaviours and structures at the highest level of complexity. Finally, Artificial Life (as a truly interdisciplinary scientific field) may be viewed as an attempt to understand high-level behaviour from low-level rules, for example, on how the simple interactions between ants and their environment lead to complex trail-following behaviour. But by far more important than studying ants itself, is to study how they organize themselves, through out a simple adaptive *mechanism* that seems to be present in many natural phenomena of our world. An understanding of such relationships in particular bio-inspired systems can suggest novel solutions to complex real-world problems such as disease prevention, pattern recognition, stock-market prediction, or data mining on the Internet (to name up a few).

One of the most well-know examples is the area of Evolutionary Computation. In the spirit of *Von Neumann*, *John Holland* has attempted to abstract the logical form of the natural process of biological evolution in what is currently known as the *Genetic Algorithm* (GA). In the GA, a genotype is represented as a character string that encodes a potential solution to a problem. For instance, the character string (*chromosome*) might encode the weight matrix of a neural network, or the rule table of any *Cellular Automata*, or in its simplest way, any pseudo-solution to any specific problem. These character strings are rendered as phenotypes via a problem-specific interpreter, which constructs, for example, the artificial neural network or the cellular automata machine specified by each genotype, evaluates its performance in the problem domain, and provides it with a specific fitness value. From this point the GA implements an artificial selection by making more copies of the character strings representing the better performing phenotypes. The GA generates variant genotypes by applying genetic operators to these character strings. The genetic operators typically consist of *reproduction*, *cross-over*, and *mutation*, with occasional usage of *inversion* and *duplication*. What is interesting is that "poor" individuals along several generations, often encode in parts of their genotypes, the key for the best solutions (artificial individuals) to become better. The best GA solution, is in some sense a product of the GA collective change of information, a product of the whole, being diversity a key aspect in the process, and a way for the artificial algorithm to balance his own exploration/exploitation duality character on the fitness landscape (space of possible solutions). Such evolutionary approaches are being applied to tasks such as optimisation, search procedures, classification, and adaptation, among others.

As the computational strategies mentioned above, *Complex dynamic systems* in general show interesting and desirable behaviours as *flexibility* (in vision or speech understanding tasks, the brain is able to cope with incorrect, ambiguous or distorted information, or even to deal with unforeseen or new situations without showing abrupt performance breakdown) or *versatibility* quoting *Dorigo* and *Colorni*, *robustness* (keep functioning even when some parts are locally damaged - *Damásio*), and they operate in a *massively parallel fashion*. As we know, systems of this kind abound in nature. A vivid example is provided by the behaviour of a society of termites (*Courtouis*). And, as a key feature, *complex dynamical systems* show and provide emergent properties. Again, this means that the behaviour of the system as a whole can no longer be viewed as a simple superposition of the individual behaviours of its elements, but rather as a side effect of their collective behaviour. Contained in this notion is the idea that properties are not *a priory* predictable from the structure of the local interactions and that they are of functional significance. The computation to be performed is contained in the dynamics of the system, which in turn is determined by the nature of the local interactions between the many elements.

Many of the dynamical computation systems that have been developed today find also their equivalent in nature, and all of them show, directly or not, important emergent properties (among other lifelike features). A non-extensive list of possible paradigmatic examples include, *Genetic Algorithms*, *Memetic Algorithms*, *Spin Glass Models*, *Connectionist Architectures* and *Artificial Neural Networks*, *Reaction-Diffusion* systems, *Self-Organizing Maps*, *Simulated Annealing* methods, *Artificial Imunne systems*, *Cellular Automata*, *L-Systems*, *Gradient Vector Flow* and *Snakes*, *Differential Evolution*, *Correlational Opponent Processing* and *Particle Swarm Optimization*.

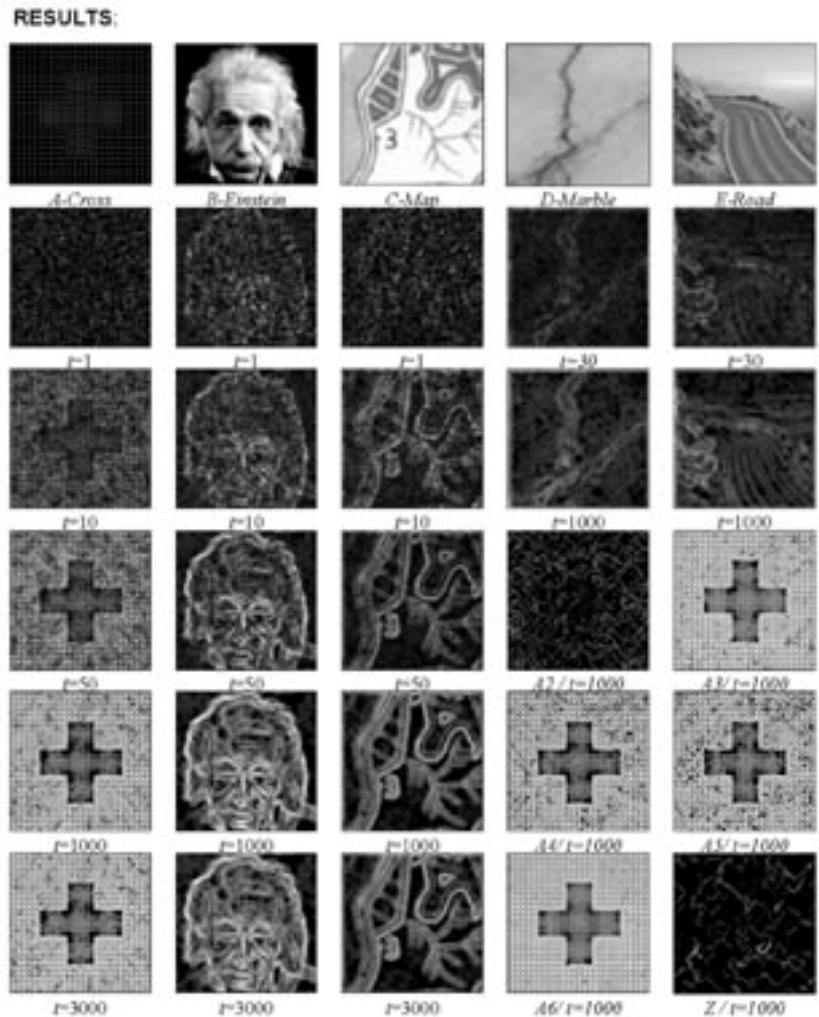

Figure 3 - Colony cognitive maps (pheromonal fields) for several iterations, on images Cross, Einstein, Map, Marble and Road. Except when indicated, parameters are those from [1]. In A2 and Z, ants are allowed to step on each other; habitats are respectively Cross and an homogeneous image. In this last case, results are similar with those found by Chialvo and Millonas [1]. A3) $k$=0.011. A4) $k$=0.019. A5) $\beta$=4.5. A6) $\beta$=2.5.

As an example, biological metaphors offer insight into many aspects of computer viruses and can inspire defences against them. That is the case with some applications of *Immunological Computation*, and *Artificial Immune Systems*. The immune system is highly distributed, highly adaptive, maintains a memory of past encounters, is self-organising in his own nature and has the ability to continually learn about new encounters. From a computational viewpoint, the immune system has much to offer by way of inspiration. Detection of specific patterns in large databases is one possible application. Autonomous alert collision systems, in route management for airplanes is another.

An important feature in many of these dynamical computational systems is that of *interaction* (e.g. competition-cooperation duality). Cooperation involves a collection of agents – global behaviours, if we strictly follow *Langton* words - that interact by communicating information, or hints (usually concerning regions to avoid or likely to contain solutions) to each other while solving a problem. This duality interaction can also be found in the well-known *Prisoner Dilemma* game theory problem, into which many Evolutionary Computation approaches are being used. The information exchanged may be incorrect at times and should alter the behaviour of the agents receiving it, yet, what emerges at the end is an robust rule in the pool of rules, which is cooperative. Another example of cooperative problem solving is the use of the Genetic Algorithm to find states of high fitness in some abstract space. In a Genetic Algorithm, members of a population of states exchange pieces of themselves or mutate to create new populations, often containing states of higher fitness. In *Artificial Neural Networks*, we can also find similar features, where the output of one neuron affects the behaviour (or state – under the light of *Cellular Automata* theory) of the neuron receiving it, and so on. Reporting to the real nature and quoting *Damásio*, we are barely beginning to address the fact that interactions among many non-contiguous brain regions probably yield highly complex biological states that are vastly more than the sum of their parts. It is important, however, to point out that the brain and mind are not a monolith: they have multiple structural levels, and the highest of those levels creates instruments or artefacts that allow for the observation of the other levels.

## MOVING ON TO THE IMPLICIT

Evolutionary Computation, whether in the form of Genetic algorithms, Genetic Programming, or Evolution Strategies, has been largely successful in solving a great variety of problems in many scientific areas. This is in part due to the fact that for these types of applications, an appropriate fitness evaluation is possible to code, and depending on the problem is a relatively simple task, that is, a performance of each solution in the population can be measured. Even if Artificial Life has made a strong rupture with the more traditional symbolic AI, by implementing non-analytical bottom-up approaches, yet and for some algorithmic paradigms like Evolutionary Computation, there is still a need for a high-level specification of purpose (or intention), a target, in order to evaluate and select solutions found so far in each generation. However, in some real-world implementations where these evaluations are hard to formalize by any group of equations, being it in the form of multiple coded constraints, specific grammars, confidence intervals or as normally by any multi-objective evaluation function, the successful and coherent application of Genetic Algorithms are jeopardize, remaining the strategy a pure random process. This is the case, in many, if not all the recent implementations of synthetic evolutionary art, or generative art and architecture. As we know, defining any aesthetic criteria is difficult, being the translation from these to an automatic set of mathematical selection rules, probably even more difficult or impossible (since, among other aspects, the relation of the art work in formation and the artist can be seen as a process of co-evolution). In the absence of any mathematical function that can map coherently the relations of, for instance, form into aesthetical value, the final result will always be a random guided-tour of some sub-space of possible and hypothetical novel solutions, not different in many aspects, to a trial and error basic process, submitted to any specific conceptual search space. This is mainly a problem of representation, since any attitude to implement those aesthetical fitness functions, mapping genotypes into the "usefulness" of the hypothetical novel forms, can be as dangerous as the objective and analytical evaluation of any final art work. This high-level mapping attitude is in itself a compression method, where the diversity of any conceptual world and the nature of its several dimensions are reduced to some aspects. *Luis Borges* words on these matters are wise: the only true map of the world, is the world itself.

There is however one way to avoid those mathematical mappings. That is of using the human observer or "artist", as an "aesthetical mapping machine", connected in real-time to the artificial evolutionary process. The first computer-graphics program where the idea was introduced was due to *Karl Sims* (1991). This program uses genetic algorithms to generate new images, or patterns, from pre-existing images. Unlike most GA systems, the selection of the "fittest" examples is not automatic, but is done by the programmer. That is, the human being selects the images which are aesthetically pleasing, or otherwise interesting, and these are used to breed the next generation, being the whole an interactive graphics environment. Although interesting in many fields, this type of Evolutionary Computation is however dependent on the human observer, and on his attitude, since the measuring of any pool of pseudo-solutions (e.g., art works in formation), at any given generation, must be evaluated and ranked by the artist looking to the monitor. For instance, there are several Genetic Algorithms, that work under this line, helping to find criminals. In order to search optimally all the combinatorial nature of the human faces, the victim points out to the GA, some aspects and features of the face of the criminal, evaluating and ranking different evolutionary proposals, generation after generation, being the final result a robot-drawing of that criminal (or at least, similar to him). In the context of artificial art, this conceptual framework has recently being followed, but still, the human-computer interaction, determines a final art work, which is far from being purely a result from an autonomous process, being the computational paradigm just a tool to achieve any artistic purpose, as a pencil still his for any *drawing* artist.

Even if those methodological approaches are interesting, there is still a profound gap on the understanding of other mechanisms that can have a crucial role on the sciences of the artificial, and predominantly on the nature of morphogenesis. As an example, for the past ten years or so, aLife research debates intensively the basic characters of those non-linear mappings between genotypes and phenotypes. What are, for instance, the necessary alphabets and their local relationships, which can promote and emerge spatial organization at the higher levels of complexity in any system? What parts and elementary mechanisms are essential, in order to have the system behaving as a *usefulness*, creative and autonomous whole? In order to find a truly innate artificial emergent life (a research field that could have an enormous impact on the synthetic computational art and architecture, itself), we must thus, study and find other approaches, which can be coupled or not with the existing aLife systems.

As a researcher in all the computational paradigms briefly described above, I believe that there are three crucial aspects in order to create truly low-level mechanisms that could autonomously emerge novel æsthetical patterns. The first aspect to discuss, is on the nature of *autonomy* itself, that is, on the nature of any autonomous mechanisms embedded in any artificial organism. Rather specifying high-level constraints, even if minimally at each chromosome fitness evaluation, we should take the opposite way, that is, to allow for any organism an implicit nature of those emerged characters. We must follow the bottom-up methodological design, till the end. The second is on the role of intrinsic *co-evolution* between parts of the artificial system, on how it can be implemented, and on his intrinsic scientific properties. By doing this, we allow the system to connect into any and possible unsupervised realm. However, and for the sake of simplicity, I will not discuss these properties on here. Finally the third aspect is related to the nature of *self-organization*, a concept that links back to autonomy and emergence.

The first aspect of autonomy involves behaviour mediated, in part, by inner mechanisms shaped by any artificial organism past experience. These mechanisms may, but need not, include explicit representations of current or future states. However, the important distinction is between a response wholly dependent on the current environmental state (given the original, "innate", bodily mechanisms), and one largely influenced by the creature's experience. The more a creature's past experience differs from that of other creatures, the more "individual" its behaviour will appear. The second aspect of autonomy, relates to know into what extend the controlling mechanisms were self-generated rather than externally imposed. That is, a distinction between behaviour which emerges as a result of self-organizing processes, and behaviour which was deliberately prefigured in the design of that organism, or organisms. This concerns for instance those behaviours that emerge, from an initial list of simple rules concerning locally interacting units, but it was neither specifically mentioned in those rules, nor foreseen when they were written. As an example, this is case of *Boids*. Ethnologists, aLife workers, and situated roboticists (e.g., *Inman Harvey*), all assume that increasingly complex hierarchical behaviour can arise in this sort of way. The more levels in the hierarchy (layers of complexity, as I like to call them), the less direct the influence of environmental stimuli will be – and the greater the behavioural autonomy. Even if, all started (or can be started, as I believe), from a simple set of environmental stimuli. A primordial soup of implicit characters. An intrinsic and kaleidoscopically genotype.

For all these reasons and within all these paradigms, we must focus our study and implement computational features that underlie the *collective* and the *distributed*, the *flexible* and the *versatile*, the *massively parallel* and the *dynamical*. Finally we should research those systems that rely on *synergy*, *cooperation* and *co-evolution*, with or without embodied evolutionary computation. One prominent example, are *Artificial Ant Systems*, a research area to which I humble contribute since 1998. In *"Godel, Escher, Bach"*, Douglas Hofstadter explores the difference between an ant colony as a whole and the individual that composes it. According to *Hofstadter*, the behaviour of the whole colony is far more sophisticated and of very different character than the behaviour of the individual ants. A colony's collective behaviour exceeds the sum of its individual member's actions (so-called *emergence*) and is most easily observed when studying their foraging activity. Most species of ants forage collectively using chemical recruitment strategies, designated by *pheromone* trails, to lead their fellow nest-mates to food sources.

This analogy with the way that real and natural ant colonies work and migrate, has suggested the definition in 1991/92 of a new computational paradigm, which is called the *Ant System* (*Dorigo / Colorni*). In these studies (again) there is *no pre-commitment* to any particular representational scheme: the desired behaviour is specified, but there is minimal specification of the mechanism required to generate that behaviour, i.e. *global behaviour evolves from the many relations of multiple simple behaviours*. Since then several studies were conducted to apply this recent paradigm – or analogous ones - in real case problems, with successful results. In short, the new heuristic has the following desirable characteristics: (1) It is *versatile*, in that it can be applied to similar versions of the same problem; (2) It is *Robust*. It can be applied with only minimal changes to other problems (e.g. combinatorial optimisation problems such as the quadratic assignment problem - QAP, travelling salesman problem - TSP, or the job-shop scheduling problem - JSP);…and (3) It is a *population based approach*. This last property is interesting since it allows the exploitation of positive feedback as a search mechanism (the collective behaviour that emerges is a form of *autocatalytic* "snow ball" - that reinforces itself - behaviour, where the more the ants follow a trail, the more attractive that trail becomes for being followed). It also makes the system amenable to parallel implementations (though, only the intrinsically parallel and distributed nature of these systems are generally considered).

One typical case of interest is that of perception. I have explored the application of Artificial Ant Systems into Pattern Recognition problems, namely to the sub-problem of image segmentation, i.e., to find homogeneous regions in any digital image, in order to extract and classify them. The application of these heuristics onto image segmentation looks very promising, since segmentation can be looked as a clustering and combinatorial problem, and the grey level image itself as a topographic map (where the image is the *ant colony playground*). Then, the distribution of the pheromone (a volatile and chemical substance) represents the memory of the recent history of the swarm, and in a sense it contain information which the individual ants are unable to hold or transmit. In this artificial system, there is no direct communication between the organisms but a type of indirect communication through the pheromonal field. In fact, ants are not allowed to have any memory and the individual's spatial knowledge is restricted to local information about the whole colony pheromone density. Particularly interesting for the present task (i.e. trying to evolve perceptive capabilities), is the fact that self-organisation of ants into a swarm and the self-organisation of neurones into a brain-like structure are similar in many respects (*Chialvo, Millonas*). Swarms of social insects construct trails and networks of regular traffic via a process of pheromone laying and following. These patterns constitute what is known in brain science as a *cognitive map*. The main differences lies in the fact that insects write their spatial memories in the environment, while the mammalian cognitive map lies inside the brain, a fact that also constitutes an important advantage in the present model. As mentioned by *Chialvo*, this analogy can be more than a poetic image, and can be further justified by a direct comparison with the neural processes associated with the construction of cognitive maps in the hippocampus. *Wilson*, for instance, forecasted the eventual appearance of what he called "a stochastic theory of mass behaviour" and asserted that "the reconstruction of mass behaviours from the behaviours of single colony members is the central problem of insect socio-biology". He forecasted that our understanding of individual insect behaviour together with the sophistication with which we will able to analyse their collective interaction would advance to the point were we would one day posses a detailed, even quantitative, understanding of how individual "probability matrices" would lead to mass action at the level of the colony. By replacing *colony members* with *neurones*, *mass behaviours* or *colony* by *brain behaviour*, and *insect socio-biology* with *brain science* the above paragraph could describe the paradigm shifts in the last twenty-five years of progress in the brain sciences.

Perception is also an important conceptual element in what can be related to autonomy. Probably a third criterion of autonomy (not listed earlier) links to the extend to which a system's inner directing mechanisms can be reflected upon, and/or selectively modified, by the individual concerned. One way in which a system can adapt its own processes, selecting the most fruitful modifications, is to use an evolutionary strategy such as the genetic algorithms mentioned above. It may be that something broadly similar goes on in human minds. But the mutations and selections carried out by GAs are modelled on biological evolution, not conscious reflection and self-modification. And it is conscious deliberation which many people assume to be the root of human autonomy. Thus, it is primarily on perception and inner recognition that the system must rely.

Moreover, perception itself, as a human feature is being modelled and analysed by *Gestalt* psychology and philosophical systems since, at least 1910 (*Wertheimer*). It is of much interest to follow that this kind of scientific works point out that perception is a product of a synergistic whole effect, i.e. the effect of perception is generated not so much by its individual elements (e.g. human neurones) as by their dynamic interrelation (collective behaviour) – phenomena that can be found easily in many computational paradigms briefly described above, or even in *Neural Network* computational models, where data generalisation, *N* dimensional matrix re-mapping, pattern classification or forecasting abilities are known to be possible. As putted by *Limin Fu* in his own words, the *intelligence* of a Neural Network emerges from the collective behaviour of neurones, each of which performs only very limited operations. Even though each individual neuron works slowly, they can still quickly find a solution by working in parallel. This fact can explain why humans can recognize a visual scene faster than a digital computer, while an individual brain cell responds much more slowly than a digital cell in a VLSI (*Very Large Scale Integration*) circuit. Also, this *brain metaphor* suggests how to build an intelligent system which can tolerate faults (fault tolerance) by distributing information redundantly. It would be easier to build a large system in which most of the components work correctly than to build a smaller system in which all components are perfect. Another feature exhibited by the brain is the associative type of memory. The brain naturally associates one thing with another. It can access information based on contents rather than on sequential addresses as in the normal digital computer. The associative, or content-addressable, memory accounts for fast information retrieval and permits partial or approximate matching. The brain seems to be good at managing fuzzy information because of the way its knowledge is represented. The key aspect is that artificial ant systems behave similarly.

Typically these systems form a structure, configuration, or pattern of physical, biological, sociological, or psychological phenomena, so integrated as to constitute a functional unit with properties not derivable from its parts in summation (i.e. non-linear) – *Gestalt* in one word (*Krippendorff*) (the English word more similar is perhaps *system*, *configuration* or *whole*). This synergetic view, derives from the holistic conviction that the whole is more than the sum of its parts and, since the *energy* in a whole cannot exceed the sum of the energies invested in each of its parts (e.g. first law of thermodynamics), that there must therefore be some quantity with respect to which the whole differs from the mere aggregate. This quantity is called synergy and in many aLife computational systems can be seen as their inherent emergent and *autocatalytic* properties (process well known in many *Reinforcement Learning* models, namely in Q-learning methods often used in autonomous-agents design (*Mitchell* / *Maes*).

Part of what we now see in these figures, was due to a model that has explored the application of these features into digital images, replacing the normal colony *habitat*, by grey levels, extending the capabilities of pheromone deposition into different situations, allowing a process of perceptual morphogenesis. In other words, from local and simple interactions to global and flexible adaptive perception. In those experiments, the emergence of network pheromone trails, for instance, are the *product* of several simple and local interactions that can evolve to complex patterns, which in some sense translate a meta-behaviour of that swarm. Moreover, the translation of one kind of low-level structure of information (present in a large number) to one meta-level is minimal. Although that behaviour is specified (and somehow constrained), there is minimal specification of the mechanism required to generate that behaviour; global behaviour evolves from the many relations of multiple simple behaviours, without global coordination, and using indirect communication (through the environment). One abstract example is the notion, within a specified population, of *common-sense*, being the meta-result a type of *collective-conscience*. Needless to say, that some features are acquired (through out the evolving relation with the habitat), being others inner components of each part. Though, what is interesting to note is that we do not need to specify them. Moreover, the present model shows important adaptive capabilities, as in the presence of sudden changes in the *habitat*. Even if the model is able to quickly adapts to one specific environment, evolving from one empty pheromonal field, *habitat* transitions point that, the whole system is able to have some memory from past environments (i.e. convergence is more difficult after *learning* and *perceiving* one *habitat*). This emerged feature of *résistance*, is somewhat present in many of the natural phenomena that we find today in our society.

In short, the design of such systems must follow a conceptual flux, where *autonomy*, *perception* and *synergy* are the key-elements. My final words are exactly about synergy within ant systems, and on how this aLife scientific essay in the intersection can help building or suggest novel 2D patterns, or even 3D architectures, as we now see on these pages. Synergy (from the Greek word *synergos*), broadly defined, refers to combined or co-operative effects produced by two or more elements (parts or individuals). The definition is often associated with the quote "the whole is greater than the sum of its parts" (Aristotle, in *Metaphysics*), even if it is more accurate to say that the functional effects produced by wholes are different from what the parts can produce alone. Synergy is a ubiquitous phenomenon in nature and human societies alike. One well know example is provided by the emergence of self-organization in social insects, via direct (mandibular, antennation, chemical or visual contact, etc) or indirect interactions. The latter types are more subtle and defined by *Grassé* as *stigmergy* to explain task coordination and regulation in the context of nest reconstruction in *Macrotermes* termites. An example, could be provided by two individuals, who interact indirectly when one of them modifies the environment and the other responds to the new environment at a later time. In other words, *stigmergy* could be defined as a typical case of environmental synergy. *Grassé* showed that the coordination and regulation of building activities do not depend on the workers themselves but are mainly achieved by the nest structure: a stimulating configuration triggers the response of a termite worker, transforming the configuration into another configuration that may trigger in turn another (possibly different) action performed by the same termite or any other worker in the colony. Another illustration of how *stigmergy* and self-organization can be combined into more subtle adaptive behaviors is recruitment in social insects. Self-organized trail laying by individual ants is a way of modifying the environment to communicate with nest mates that follow such trails.

It appears that task performance by some workers decreases the need for more task performance: for instance, nest cleaning by some workers reduces the need for nest cleaning. Therefore, nest mates communicate to other nest mates by modifying the environment (cleaning the nest), and nest mates respond to the modified environment (by not engaging in nest cleaning); that is *stigmergy*.

In other words, perception and action only by themselves can evolve adaptive and flexible problem-solving mechanisms, or emerge communication among many parts. The whole and their relationships (that is, the next layer in complexity) emerges from the relationship of many parts, even if these latter are acting strictly within and according to any sub-level of basic and simple strategies, *ad-infinitum* repeated. Quoting *Einstein*, the system "should be made as simple as possible, but not simpler".

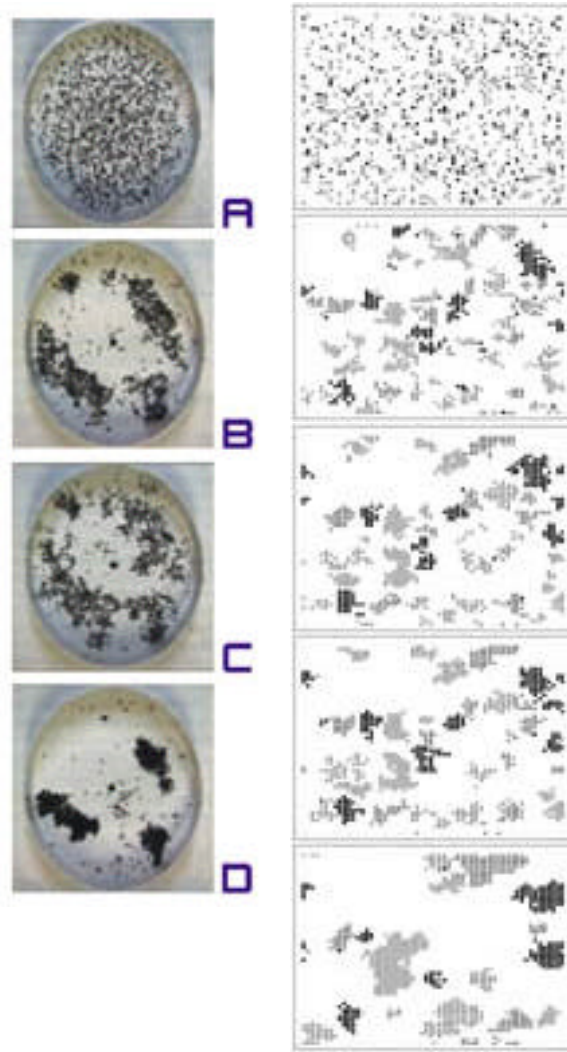

On the rigth, a sequential clustering task of corpses performed by a real ant colony. 1500 corpses are randomly located in a circular arena with radius = 25 cm, where *Messor Sancta* workers are present. The figure shows the initial state (a), 2 hours (b), 6 hours (c) and 26 hours (d) after the beginning of the experiment. On the left, an artificial swarm evolving clusters of semantically similar data items (*Ramos*, 2001).

Division of labor is another paradigmatic phenomena of *stigmergy*. Simultaneous task performance (parallelism) by specialized workers is believed to be more efficient than sequential task performance by unspecialized workers. Parallelism avoids task switching, which costs energy and time. A key feature of division of labor is its plasticity. Division of labor is rarely rigid. The ratios of workers performing the different tasks that maintain the colony's viability and reproductive success can vary in response to internal perturbations or external challenges.

But by far more crucial to the design of any collective pattern artificial system, is how ants form piles of items such as dead bodies (corpses), larvae, or grains of sand. There again, *stigmergy* is at work: ants deposit items at initially random locations. When other ants perceive deposited items, they are stimulated to deposit items next to them, being this type of cemetery clustering organization and brood sorting a type of self-organization and adaptive behavior. *Théraulaz* and *Bonabeau* described for instance, a model of nest building in wasps, in which wasp-like agents are stimulated to deposit bricks when they encounter specific configurations of bricks: depositing a brick modifies the environment and hence the stimulatory field of other agents. These asynchronous automata (designed by an ensemble of

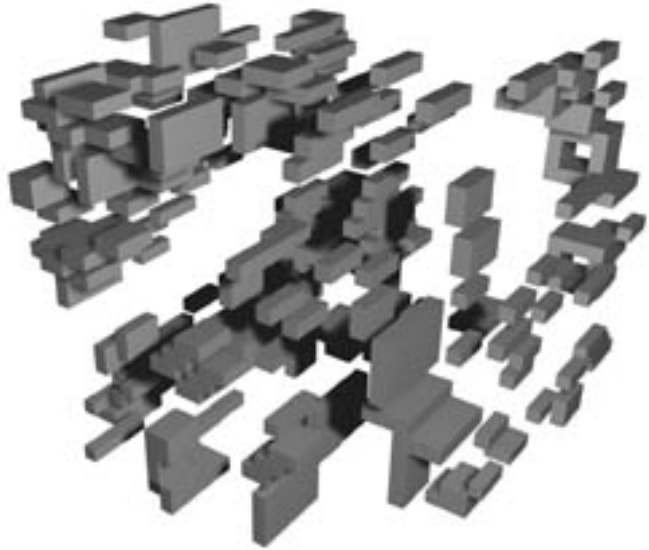

A swarm emerging global 3D patterns, from multiple local configuration stimulus.

algorithms) move in a 3D discrete space and behave locally in space and time on a pure stimulus-response basis. There are other types of examples (e.g. prey collectively transport), yet *stimergy* is also present: ants change the perceived environment of other ants (their cognitive map, according to *Chialvo* and *Millonas*), and in every example, the environment serves as medium of communication.

What all these examples have in common is that they show how *stigmergy* can easily be made operational. As mentioned by *Bonabeau*, that is a promising first step to design groups of artificial agents which solve problems: replacing coordination (and possible some hierarchy) through direct communications by indirect interactions is appealing if one wishes to design simple agents and reduce communication among agents. Another feature shared by several of the examples is incremental construction: for instance, termites make use of what other termites have constructed to contribute their own piece. In the context of optimization (though not used on the present works), incremental improvement is widely used: a new solution is constructed from previous solutions (see ACO paradigm, *Dorigo et al*). Finally, *stigmergy* is often associated with flexibility: when the environment changes because of an external perturbation, the insects respond *appropriately* to that perturbation, as if it were a modification of the environment caused by the colony's activities. In other words, the colony can collectively respond to the perturbation with individuals exhibiting the same behavior. When it comes to artificial agents, this type of flexibility is priceless: it means that the agents can respond to a perturbation without being reprogrammed in its intrinsic features to deal with that particular instability. The system organizes itself in order to deal with new *object* classes (conceptual ideas translated to the computer in the form of basic 2D/3D forms), or even new sub-classes. This task can be performed in real time, and in robust ways due to system's redundancy.

Data and information clustering is one of those problems in which real ants can suggest very interesting heuristics for computer scientists, and it is in fact a classic strategy often used in Image and Signal Processing. For the past two years, I have been developing research on these areas. Many experiments are now under their way at the CVRM-IST Lab (for instance, real-time marble and granite image classification, image and data retrieval, etc), along with the application of Genetic Algorithms, Neural Networks, and many others (based strictly on natural computation paradigms) into many problems in Natural Resources Management, like forecasting water quality and control on river networks. Surprisingly, these studies can help us to understand how artificial stigmergic systems can be implemented in order to produce novel and autonomous patterns.

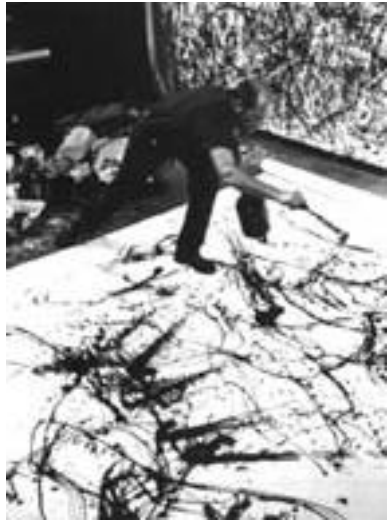

Jackson Pollock

These algorithms mimic the inner stimuli response threshold functions of each organism composing the system, and in some sort, what happens in several species of ants, where workers have been reported to sort their larvae or form piles of corpses – literally cemeteries – to clean up their nests. *Chrétien* has performed experiments with the ant *Lasius niger* to study the organization of cemeteries. Other experiments include the ants *Pheidole pallidula* reported by *Denebourg*, and many species actually organize a cemetery. If corpses, or more precisely, sufficiently large parts of corpses are randomly distributed in space at the beginning of the experiment, the workers form cemetery clusters within a few hours, following a behavior similar to aggregation. If the experimental arena is not sufficiently large, or if it contains spatial heterogeneities, the clusters will be formed along the edges of the arena or, more generally, following the heterogeneities. The basic mechanism underlying this type of aggregation phenomenon is an attraction between dead items mediated by the ant workers: small clusters of items grow by attracting workers to deposit more items. It is this positive and autocatalytic feedback that leads to the formation of larger an larger clusters. In this case, it is therefore the distribution of the clusters in the environment that plays the role of stigmergic variable.

Finally, the simulated ecology of different stimuli response threshold organisms, triggered by the seeds of these stigmergic processes, whether in the form of 3D local configurations, or by the qualitative values of any conceptual data items, must not be overestimated. Above all, the behaviour that emerges from all these spatial-temporal relationships conduct us into the realm of what nature is about: dynamical patterns of complexity. Not chaotic or purely rendered at random, but at the edge of chaos (*Langton*), where *creative* and autonomous aLife survives. As reported recently by *Nature* magazine (Sept., 13, 2000), research suggests that the abstract works of artists such as *Jackson Pollock* are esthetically pleasing because they obey fractal rules similar to those found on the natural world. *Pollock* was known to have swung his paint back and forth like a pendulum, using a can on the end of a string with a hole punched in it. Researchers (*Jensen*) have found that if a swinging pendulum is hit with a hammer at just the right frequency (slightly less than the natural rhythm of the pendulum), its motion becomes chaotic and the paint traces out very convincing "fake *Pollocks*". However, the artist had no idea of fractals or chaotic motion. This seems to be in line with the actual synthetically computational art, where there is a need to reference some kind of *external* artifact or mechanism, but nevertheless and as it appears, not those of the self whether they are conscious, unconscious, intuitive or not. Synthetically generative art, and above all, artificial systems of morphogenesis of any kind, should be much more about what scientists call "complexity", and rely on nature as a physical generative force of ontological significance. Moving on to the implicit, rather on the specific.

FURTHER READING MATERIAL:

[http://www.lxxl.pt/aswarm/aswarm.html]:

Leonel Moura, **Swarm Paintings – Non-Human Art**, in ARCHITOPIA Book / Catalogue, Art, Architecture and Science, J.L. Maubant and L. Moura (Eds.), pp. 1-24, Ministério da Ciência e Tecnologia, Feb. 2002.

[http://alfa.ist.utl.pt/~cvrm/staff/vramos]:

[ ] Vitorino Ramos, Filipe Almeida, **Artificial Ant Colonies in Digital Image Habitats - A Mass Behaviour Effect Study on Pattern Recognition**, ANTS'2000 – 2$^{nd}$ *Int. Workshop on Ant Algorithms (From Ant Colonies to Artificial Ants)*, M. Dorigo, M. Middendorf & T. Stüzle (Eds.), Brussels, Belgium, 7-9 Sep. 2000.
[ ] Vitorino Ramos, **The Biological Concept of Neoteny in Evolutionary Colour Image Segmentation - Simple Experiments in Simple Non-Memetic Genetic Algorithms**, in *Applications of Evolutionary Computation*, E.J.W. Boers et al. (Eds.), SPRINGER-VERLAG, 2001.
[ ] Vitorino Ramos, Fernando Muge, **Map Segmentation by Colour Cube Genetic K-Mean Clustering**, in *Research and Advanced Technology for Digital Libraries*, J. Borbinha and T. Baker (Eds.), SPRINGER-VERLAG, 2000.
[ ] Vitorino Ramos, **Self-Organized Data and Image Retrieval as a Consequence of Inter-Dynamic Synergistic Relationships in Artificial Ant Colonies**, ICEIS´02 – 4$^{th}$ *Int. Conf. on Enterprise Information Systems*, Ciudad-Real, Spain, 3-6 April, 2002.
[ ] Vitorino Ramos, Juan J. Merelo, **Self-Organized Stigmergic Document Maps: Environment as a Mechanism for Context Learning**, AEB´2002 – 1$^{st}$ *Spanish Conference on Evolutionary and Bio-Inspired Algorithms*, Mérida, Spain, 6-8 Feb. 2002.
[ ] Vitorino Ramos, Leonel Moura, **The MC2 Project [Machines of Collective Conscience]**, in the Official Newspaper of the *UTOPIA Biennial Art Exposition*, Cascais, Portugal, July 12-22, 2001.
[ ] Vitorino Ramos, Luis Ribeiro, **Exploring Temporal Rule-Relations in Water Quality Time Series with Swarm Intelligence**, AIS´2002 - *Artificial Intelligence, Simulation and Planning in High Autonomy Systems*, Lisbon, Portugal, April 7-10, 2002.
[ ] Vitorino Ramos, Pedro Pina, Fernando Muge, **Mining Textural Features with Artificial Ant Colonies – Towards a Dynamic and Continuous Classification of Polished Marble Slabs**, RecPad'2002, 12$^{th}$ *Int. Portuguese Conf. On Pattern Recognition*, Aveiro, Portugal, 2002.
[ ] Vitorino Ramos, **Stigmergic Design: Collective Patterns in 3D-Space and Time**, DESIGN & NATURE 2002, WIT Press, Udine, Italy, 2002.
[ ] Vitorino Ramos, **From Biological Models of Stigmergy in Social Insects to Collective, Dynamical and Self-Organized Data Mining Systems**, DATA MINING 2002, WIT Press, Bologna, Italy, 2002.